\documentclass{article}

\usepackage{arxiv}

\usepackage[utf8]{inputenc} %
\usepackage[T1]{fontenc}    %
\usepackage{hyperref}       %
\usepackage{url}            %
\usepackage{booktabs}       %
\usepackage{amsfonts}       %
\usepackage{nicefrac}       %
\usepackage{microtype}      %
\usepackage{lipsum}         %
\usepackage{graphicx}
\usepackage[square,sort,comma,numbers]{natbib}
\usepackage{doi}

\usepackage{amsmath}
\usepackage{bbm}

\usepackage{comment}
\usepackage[utf8]{inputenc} %
\usepackage[T1]{fontenc}    %
\usepackage{hyperref}       %
\usepackage{url}            %
\usepackage{booktabs}       %
\usepackage{amsfonts}       %
\usepackage{nicefrac}       %
\usepackage{microtype}      %
\usepackage{xcolor}         %
\usepackage{multirow}
\usepackage{array}
\newcolumntype{P}[1]{>{\centering\arraybackslash}p{#1}}
\newcolumntype{M}[1]{>{\centering\arraybackslash}m{#1}}
\usepackage[most]{tcolorbox}
\usepackage{float}
\usepackage{placeins}

\makeatletter
\def\moverlay{\mathpalette\mov@rlay}
\def\mov@rlay#1#2{\leavevmode\vtop{%
   \baselineskip\z@skip \lineskiplimit-\maxdimen
   \ialign{\hfil$\m@th#1##$\hfil\cr#2\crcr}}}
\newcommand{\charfusion}[3][\mathord]{
    #1{\ifx#1\mathop\vphantom{#2}\fi
        \mathpalette\mov@rlay{#2\cr#3}
      }
    \ifx#1\mathop\expandafter\displaylimits\fi}
\makeatother

\date{}
\title{Dataset Summarization by K Principal Concepts}
\usepackage{todonotes} %

\author{%
  Niv Cohen  $\quad$ Yedid Hoshen%
  \\
  School of Computer Science and Engineering\\
  The Hebrew University of Jerusalem, Israel\\
}

\hypersetup{
pdftitle={A template for the arxiv style},
pdfsubject={q-bio.NC, q-bio.QM},
pdfauthor={David S.~Hippocampus, Elias D.~Striatum},
pdfkeywords={First keyword, Second Disentanglement, More},
}

\begin{document}

\maketitle

\begin{abstract}
 
We propose the new task of $K$ principal concept identification for dataset summarizarion. The objective is to find a set of $K$ concepts that best explain the variation within the dataset. Concepts are high-level human interpretable terms such as ``tiger'', ``kayaking'' or ``happy''. The $K$ concepts are selected from a (potentially long) input list of candidates, which we denote the \textit{concept-bank}. The concept-bank may be taken from a generic dictionary or constructed by task-specific prior knowledge. An image-language embedding method (e.g. CLIP) is used to map the images and the concept-bank into a shared feature space. To select the $K$ concepts that best explain the data, we formulate our problem as a $K$-uncapacitated facility location problem. An efficient optimization technique is used to scale the local search algorithm to very large concept-banks. The output of our method is a set of $K$ principal concepts that summarize the dataset. Our approach provides a more explicit summary in comparison to selecting $K$ representative images, which are often ambiguous. As a further application of our method, the $K$ principal concepts can be used to classify the dataset into $K$ groups. Extensive experiments demonstrate the efficacy of our approach.    
\end{abstract}

\section{Introduction}

\begin{figure*}[h]
\centering
\includegraphics[width=16cm]{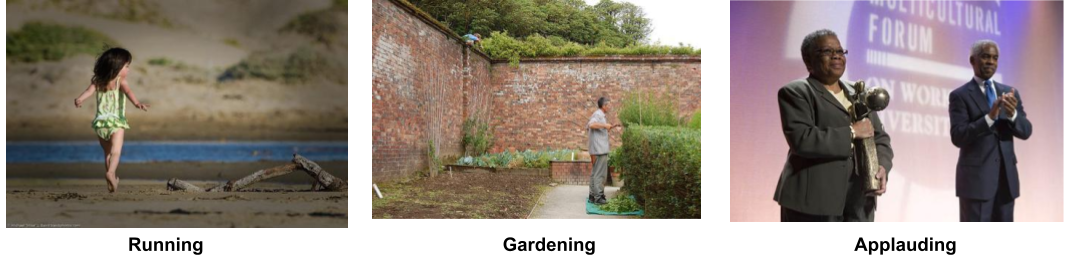}
\caption{ Summarizing a dataset (e.g. \textit{Stanford Activity} \cite{yao2011human}) with $K$ representative images can be ambiguous. Identifying the key concepts in the data (e.g `running', `applauding', `gardening') is often clearer.
}
\label{fig:concepts}

\end{figure*}

Summarizing a large image dataset into a small number of words is an important but challenging research task. This is particularly helpful for large photo storage service providers such as Google Photos or Instagram, each possessing more than a billion photo albums. Such summarization provides fast, but relatively accurate insights into dataset content \cite{singh2019image,kim2014joint}. Here we propose a new setting, focusing on an extreme form of summarization, finding a short list of $K$ concepts that retain the maximal information on the variation in the dataset. As an example, let us assume a photo collection containing humans performing various activities. A possible summary of the dataset can be a set of the $K$ most prominent featured activities (Fig.\ref{fig:concepts}). For example, for $K=3$ the principal concepts may be \{`running', `applauding', `gardening'\}.

Our approach first requires a long list of potentially valid concepts, which can potentially describe groups in the dataset. This long list, that we name the \textit{concept-bank}, can be generic (e.g. all the nouns in the WordNet database \cite{miller1995wordnet}). Alternatively, the valid concepts may be more restricted, encoding prior knowledge about the desired results. Aiming to find concepts corresponding to activities, colors or musical instruments, we may use a concept-bank restricted to such attributes. Given the concept-bank, the task is to select the $K$ principal concepts which provide the most informative description of the variation in the dataset. Although creating a concept-bank may initially appear to be a daunting task, we find that it can be performed by a very fast, semi-automatic process. 

To select $K$ principal concepts, we first use an image-language model (such as CLIP \cite{radford2021learning}) to embed all the concepts in the concept-bank and all the images in the dataset into a common embedding space. The objective becomes the selection of $K$ concepts such that the sum of distances between every image and its nearest selected concept is minimal. While this task is reminiscent of $K$-means, it is in fact different, as the search space for the centers is discrete while in $K$ means the space is unconstrained. Instead, this task is an instance of the unconstrained $K$ facility location problem, a well studied \cite{cornuejols1983uncapicitated}, NP-hard algorithmic problem. Although advanced methods exist for the solution with approximation guarantees, the most popular methods do not scale to our task. Instead, we suggest solving the optimization task using a more scalable approach that can be seen as a discretized version of $K$-means.

As a downstream application, our method can be used for conceptual grouping within image datasets. Given the $K$ principal concepts generated in the previous stage, a zero-shot learning approach may be used to classify each image into one (and only one) of the $K$ concepts. One advantage of this approach is the ability to guide the grouping by selection of the concept-bank. The grouping of images by activity, for instance, can be easily accomplished by selecting a concept-bank only containing descriptions of activities, forcing the images to be grouped accordingly. 

Most related to this task are zero-shot classification \cite{radford2021learning} and unsupervised clustering \cite{van2020scan}. Zero-shot classification is able to discover the concept to which each image is most related, but it does not by itself discover the $K$ principal concepts for a dataset given a much longer list of allowed concepts. While some greedy heuristics may be attempted, they do not perform well for large concept-banks, and due to correlations may also choose overlapping concepts (e.g. `dog' and `poodle'). Alternatively, unsupervised clustering is limited only to visual similarity and requires ad-hoc inductive biases to obtain meaningful groups.  

We evaluate our method on $3$ standard object classification datasets, and $4$ more complex datasets with non-standard groupings, demonstrating our $K$ principal concept identification and grouping capabilities. We show that our method is able to identify principal concepts close to the ground truth class-names. We further show that concept-list guidance is necessary for achieving accurate grouping results on non-standard datasets and can also increase accuracy on standard datasets. 

Our main contributions are: 

\begin{enumerate}

\item Introducing the task of $K$ principal concept identification for dataset summarization.

\item Reducing our objective to the well-studied facility location problem, and suggesting a scalable and effective solution for the optimization task.

\item Proposing pre- and post-processing methods for selecting concepts with the appropriate level of summarization.

\item Demonstrating that the identified principal concepts are effective for image grouping and allow better guidance for the grouping process. 

\end{enumerate}

\section{Related Work}
\label{sec:related_works}

\textbf{Summarizing visual datasets.}
Video or image dataset summarization is an important task that has typically been addressed by presenting several representative images. Video summarization typically identifies key frames that capture the main themes in the video \cite{potapov2014category,lee2012discovering,kim2014joint}. Image summarization have been addressed using a range of approaches, including generative modelling \cite{singh2019image}, story graphs \cite{celikkale2021generating} and multi-modal data \cite{zhu2021graph}.

\textbf{Concept Based Learning.}  Concept Bottleneck Models \cite{koh2020concept,chen2020concept,losch2019interpretability} use a network to find an intermediate set of human-specified concepts, allowing better interpretability. Another line of work uses language cues to localize visual objects within the image \cite{plummer2017phrase,kong2014you,wang2016structured}. These works aim to identify concepts within images, rather than identifying categories across the dataset. %

\textbf{Joint embedding for images and text.} 
A key motivation for looking into joint embedding is reducing the requirement for image annotations \cite{mori1999image} \cite{quattoni2007learning,joulin2016learning,sariyildiz2020learning}. %
\cite{radford2021learning} presented a new method, CLIP, that maps images and sentences into a common space. %
Our method relies on the infrastructure provided by CLIP, but does not assume the set of image names is provided. CLIP was recently followed by a line of similar works \cite{zhai2022lit,li2022masked,jia2021scaling}. As these models are rapidly developing, we use CLIP as our backbone, but results can be easily adapted to utilize any similar backbone model. %

\textbf{Uncapacitated facility location problem (UFLP).} The UFLP \cite{kuehn1963heuristic,guha1999greedy} problem is a long-studied task in economics, computer science, operations research and discrete optimization. It aims to open a set of facilities, so that they serve all clients at a minimal cost. %
Different solutions methodologies have been applied to the task including: greedy methods \cite{arya2004local}, linear-programming with rounding \cite{shmoys1997approximation} and linear-programming primal-dual methods \cite{jain2001approximation}. Here, we are concerned with the Uncapacitated K-Facility Location Problem (UKFLP) \cite{cornuejols1983uncapicitated,jain2002new}, which limits the number of facilities to $K$. We formulate our optimization objective as the UKFLP and use a fast, relaxed variant of the Local Search method \cite{arya2004local}.

\begin{figure*}[h]
\centering
\includegraphics[width=16cm]{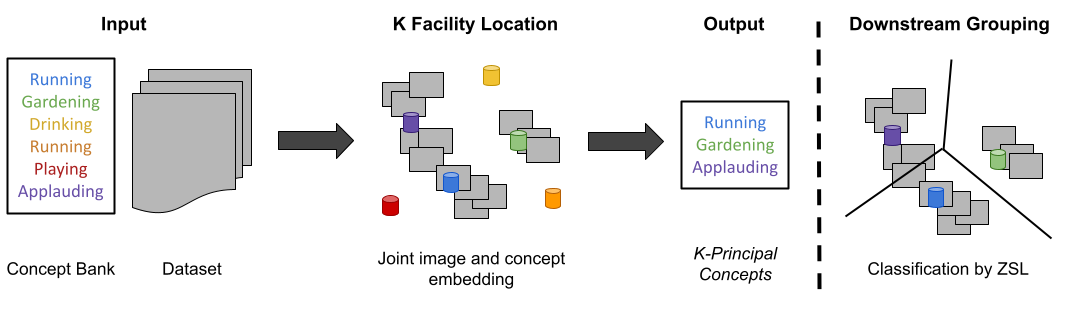}
\caption{ System diagram. Our method takes as input an image dataset and a list of concepts (here a list of activities). The images and concepts are embedded into a joint space using CLIP. A $K$ uncaptacitated facility location method is used to retrieve the $K$ principal concept. Optionally, the $K$ principal concepts can be used to classify the dataset into $K$ groups using standard zero-shot classification methods.   
}
\label{fig:method}

\end{figure*}

\textbf{Image Grouping.} Deep features trained using self-supervised criteria are extensively used for image grouping e.g. \cite{xie2016unsupervised,yang2017towards}. More recent approaches directly optimize clustering objectives during feature learning e.g. \cite{caron2018deep,chang2017deep,haeusser2018associative}. Choosing features that align with human semantics requires inductive bias. A promising line of approaches use carefully selected augmentations to remove the nuisance attributes and direct learning towards more semantic features   \cite{wu2019deep,niu2020gatcluster,shiran2019multi, tsai2021mice}. The work by Van Gansbeke et al. \cite{van2020scan} suggested a two stage approach, where features are first learned using a self-supervised task, and then used as a prior for learning the features for grouping. In practice, we often look for groups which would be balanced in size, at least approximately. Many works utilize an information theoretic criterion to impose such balancing \cite{hu2017learning,ji2019invariant,darlow2020dhog}. %
Other work uses partial supervisory information to infer groups \cite{han2019learning,finley2005supervised, guerin2021combining,shen2021structure}.

\textbf{Multi Modal Image Grouping.} Approaches such as multi-modal clustering \cite{jiang2019dm2c,jin2015cross} address the tasks of grouping data, where the same kind of samples can be represented with different data modalities. We note that these interesting tasks significantly differ from ours: our additional modality (text) describes the set of possible groups, rather than additional samples given in a new modality. Other prior works, which are more similar to our approach include color quantization using colors that are named in the English language. In these works one divides all colors into a discrete number of color groups \cite{van2009learning,yu2018beyond,mojsilovic2005computational}. Color name-based identification was further applied to other tasks, such as image classification, visual tracking, action recognition \cite{van2015overview} and person identification \cite{yang2014salient, prates2016predominant}. Our approach can be seen as extending these ideas from pixel color to whole images. 

\section{K Principal Concept Identification}

We propose the novel task of $K$ principal concept identification for dataset summarization. The goal is to select $K$ concepts from a (potentially very long) concept-bank that best describe the variation in data (Fig.\ref{fig:method}). We describe the concept-bank creation process in Sec.~\ref{sec:word_list}. Given a concept-bank, the selection of $K$ concepts corresponding to the group names is described in Sec.~\ref{subsec:method:formulation}-\ref{subsec:method:group}.  Having selected the $K$ principal concepts, images can be assigned to groups using standard zero-shot clustering, and accuracy can be further improved by an adapter network (Sec.~\ref{sec:post_proc}).

\subsection{Concept-bank Creation}\label{sec:word_list}
Our method requires a list of allowed concepts. Although the list can be generated manually, this can be tedious. We suggest using a semi-automatic way of creating the list. First, the operator provides a single word (or phrase) describing the concept-bank they wish to group the images by. Examples of such attributes include: ``activities'', ``objects'', ``dogs'' or ``musical instruments''. A concept-bank is then retrieved from an online Word Bank such as \cite{enchanted2008wor}. We note that even a very general list such as WordNet nouns (for ``objects'') containing over $82k$ can be very effective for a large variety of datasets as demonstrated in Sec.\ref{sec:exp}. However, there is an advantage in more fine-grained guidance. If we wish to avoid identifying specific concept categories (e.g. ``musical instruments''), such concepts should \textit{not} be included in the concept-bank. By specifying just the desired concept class, we can limit the type of possible image groupings, and ambiguities can be avoided.

\subsection{K Principal Concept Identification Formulation}
\label{subsec:method:formulation}

Our goal is to discover the $K$ concepts from the concept-bank that best describe the variation in the image dataset. We are given $N_I$ images, which are mapped into feature vectors $\{v_1 \ldots v_{N_I}\}, v_i \in \mathbb{R}^d$.
A concept-bank consisting of $N_W$ concepts describing possible image groupings is provided by the process explained in Sec.\ref{sec:word_list}. Every concept is mapped into a vector embedding $\{u_1 \ldots u_{N_W}\}, u_i \in \mathbb{R}^d$, the set of all concept embeddings is denoted as ${\mathcal{W}}$. 
We aim to select $K$ principal concepts that best describe the variation in the data. We denote the principal concepts by a corresponding set of vectors $\{p_1 .. p_K\}, p_k \in {\mathcal{W}}$. Each image is assigned to its nearest principal concept, resulting in $K$ image groups $\{S_1 .. S_K\}$. Therefore, each group consists of conceptually homogeneous images. %

\subsection{Removing Overly General Concepts}
\label{subsec:method:filter}
Although we assume all plausible concepts are contained in the concept-bank ${\mathcal{W}}$, some concepts may have a meaning that is too general. Such concepts may not explain much of the dataset variation and may even be related to all the images in the dataset. Examples for such concept are: `entity', `abstraction', `thing', `object', `whole'.  Therefore, we wish to filter such uninformative concepts out of our list.

To remove such concepts we rely on the following intuition: concepts which are very general can describe many other concepts in the list. We therefore measure how well a concept describes other concepts using the inner product between the normalized embeddings. To filter such uninformative concepts, we look for concepts with high average correlation to other concepts in the list. We first calculate the average embedding of all concepts in the list:
 \begin{equation}
    \label{eq:avg_noun}
 u_{avg} = \frac{1}{N_W} \sum_{i = 1..N_W} u_i
 \end{equation}
 We than calculate the generality score $s$ for each concept, as the inner product between its embedding $u_i$ and the average concept embedding $u_{avg}$:
  \begin{equation}
    \label{eq:generality_score}
 s(u_{i}) = u_i \cdot  u_{avg}
 \end{equation}
We find that this score is indeed higher for the less specific concepts described earlier. We remove from the list all concepts that have a ``generality score'' $s$ higher than some quantile level $0 < q \leq 1$, and define the new sublist ${\mathcal{W}}_q \subseteq {\mathcal{W}}$ ($|{\mathcal{W}}_q| \approx q\cdot |{\mathcal{W}}|$, where $|.|$ denotes the length of a set). 

We propose an unsupervised distributional criterion for choosing the quantile $q$ for each dataset. We first run our method on a set of values of $q = [0,0.05,0.1...1]$. For each value of $q$, we obtain concept assignments and calculate the distributional entropy. We select the $q$ for which the principal concepts $W_q$ form the most balanced grouping i.e., with the highest entropy. See Sec.\ref{sec:analysis} for an ablation. 

\subsection{Grouping with the K Principal Concepts}
\label{subsec:method:group}
We consider a group of images $S_k$ describable by a single concept $p_k$ if the embeddings of its associated images are near the embedding of the concept $p_k \in {\mathcal{W}}_q$. We formulate this objective, using the within-group sum of squares (WCSS) loss:

\begin{equation}
\label{eq:wcss}
\begin{array}{rrclcl}
\displaystyle \min_{\{p_1 .. p_K\},\{S_1..S_K\}} & \sum_{k=1}^K \sum_{v \in S_k} \left\Vert v - p_k \right\Vert^2\\
\textrm{s.t.} 
& p_k \in   {\mathcal{W}}_q  \\ 
\end{array}
\end{equation}

The objective is to find assignments $\{S_1..S_k\}$ and principal concepts $\{p_1..p_k\} \subseteq {\mathcal{W}}_q$, such that the sum of square distances for images and the assigned concept is minimal. Note that this is different from $K$-means as the group centers are constrained to the discrete set of concepts $\mathcal{W}$ whereas in $K$-means they are unconstrained. An efficient optimization method is proposed in Sec.~\ref{sec:optimization}.

\subsection{Post-processing}
\label{sec:post_proc}
\textbf{Moving up the WordNet hierarchy}
When the initial concept-bank is as long as the entire WordNet dictionary, we aim to replace too specific principal concepts (e.g. `bulbul') with ones that better describe the entire group of images (e.g. `bird'). Our dataset was already grouped based on the semantics of the concept-bank $\mathcal{W}_q$ by obtaining $K$ principal concepts. For each group $\{S_1..S_k\}$, we retrieve the $M$ concepts that are nearest to the group center. All the $M$ concepts are very close to the center of the group because M is chosen as a small number ($M=50$) when compared to the original concept-bank length ($|\mathcal{W}|=82k$). We then select the most abstract concept out of the top $M$ concepts.  In order to measure the abstraction level, we count the number of concepts below the examined concept in the WordNet hierarchy.

\begin{figure}[t]
\centering
\includegraphics[width=8cm]{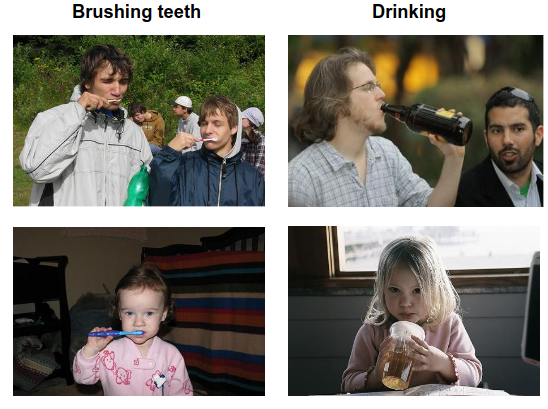}
\caption{Four images selected from the Stanford Activity dataset. The images can be grouped by coarse-object category (`adult humans', `toddler') or by activity (`drinking', `brushing teeth'). %
Our method uses a concept-bank to determine the concepts that are allowed for grouping  (in this case activities: `walking', `drinking', `brushing teeth' etc.). 
}
\label{fig:stan_act}

\end{figure}

\textbf{Tip Adapter} While direct application of zero-shot classification on the $K$ principle concepts can already achieve high grouping accuracy, better results can be obtained by fine-tuning the discovered groupings using the Tip Adapter \cite{zhang2021tip}. Further details can be found in the appendix.

\section{Optimization}
\label{sec:optimization}

\begin{table*}[t]
\begin{center}
\caption{Glossary of  CIFAR10 class names and assigned concepts for our method and ZS-Naive (ZSN). PP notes the post-processed version of the concepts. Full table can be found in the appendix. %
}
\label{table:cifar10_nouns}
\begin{tabular}{cccccccccc}
\toprule

\textbf{Ground Truth} &	horse	&		frog	&	bird	&	dog		&	automobile	&	ship	&	truck \\
\toprule
 \textbf{Ours} &	chukker  		&	southwestern toad  	&	policeman bird  	&	maltese dog  	&		Jowett  	&	pilot boat  	&	milk float   \\

  \textbf{PP} &	equine		&	frog	&	bird	&	dog		&	vehicle	&	ship	&	vehicle \\

\midrule
 \textbf{ZSN} &	chukker	&	vaulting horse	&	Seattle Slew	&	bumper car	&	tachograph	&	Jowett	&	milk float  \\
\textbf{PP} &	horse	&	horse	&	ungulate	&	driver	 &	container	&	fastness	&	truck \\

\bottomrule
\end{tabular}
\end{center}
\end{table*}

\subsection{The Uncapacitated Facility Location Problem}
We formalize our optimization problem, by restating it as an uncapacitated K-facility location problem (UKFLP). The UKFLP is a long studied discrete optimization task (see Sec.~\ref{sec:related_works}). In the UKFLP task we are asked to ``open'' $K$ ``facilities'' out of a larger set of sites ${\mathcal{W}}_q$, and assign each ``client'' to one of the $K$ facilities, such that the sum of distances between the ``clients'' and their assigned ``facilities'' is minimal. In our case, the clients are the image embeddings $v_1,v_2..v_{N_I}$, which are assigned to a set of $K$ concept embeddings selected from the complete concept-bank ${\mathcal{W}}_q$. We look to optimize an assignment variable $x_{ij} \in \{0,1\}$ indicating whether the ``client'' $v_i$ is assigned to the ``facility'' $u_j$. We also use a variable $y_{j} \in \{0,1\}$ to determine if a facility was opened in site $j$ (if the concept $u_j$ is the center of a group). The optimal assignment should minimize the sum squared distance between each image and its assigned concept. The squared distance between image $v_i$ and concept $u_j$ is denoted $d_{ij}$. We can now restate our loss as:

\begin{equation}
\begin{array}{rrclcl}
\displaystyle \min_{x_{ij}, y_j} & \sum_{i\in 1..N, j \in 1..N_W} d_{ij}x_{ij}\\
\textrm{s.t.} & \forall i\in 1..N : \sum_{j\in 1..N_W}x_{ij} = 1  \\
& x_{ij} \leq y_j   \\
& \sum_{j \in 1..N_W} y_j\leq K   \\
\end{array}
\end{equation}

Where the bottom two constraints limit the number of concepts to at most $K$. Solving UKFLP is NP-hard, and the problem of approximation algorithms for UKFLP have been studied extensively both in terms of complexity and approximation ratio guarantees (see Sec.\ref{sec:related_works}). Yet, as the distance matrix $d_{ij}$ is very large, we could not run the existing solutions at the scale of many datasets (e.g. there may be as many as $82k$ concepts-``facilities'' and a few hundred thousands images-``clients''). We therefore suggest a relaxed version of the popular Local Search algorithm.

\subsection{Local Search algorithm}
The Local Search algorithm \cite{arya2004local} is an effective, established method for solving facility location problems. Instead of looking for the optimal assignment at once, it looks for swaps between open and closed facilities that decrease the loss. It starts with ``forward greedy'' initialization: in the first $K$ steps, we open the new facility (choose a new concept embedding as a group center) that minimizes the loss the most, among all unopened sites (unselected concepts). After initialization, we iteratively perform the following procedure: In each step, we look to swap $r$ of our selected concepts by $r$ unselected concepts, such that the loss is decreased. If such concepts are found, the swap is applied. We repeat this step until better swaps cannot be found or the maximal number of iterations is reached, making it slow to run even for a small dataset. With $r=1$  this is also known as the Partitioning Around Medoids (PAM) algorithm.

\subsection{Local Search Location Relaxation Method}
\label{sec:local_relaxation}
As our task is very high-dimensional, running Local Search (or similar UKFLP algorithms) becomes too slow to be practical. Therefore, we suggest an alternative inspired by the Expectation Minimization (EM) algorithm for $K$-means \cite{moon1996expectation}. We look for concepts $\{p_1 .. p_K \}$ nearest to the center of each group. This relaxation approach of searching concepts in continuous Euclidean space is much faster to compute (with complexity $O(N_I + N_W)$). Our method is initialized with a set of centers $\{c_1 .. c_K \}$ and iterates the following steps until convergence: (i) We assign each of our images $v_1..v_N$ to groups $\{S_1..S_K\}$ according to the nearest group center (``Voronoi tessellation'')

\begin{equation}
    S_{k'} = \{{v_i | \left\Vert  v_i - c_{k'} \right\Vert^2  \leq \left\Vert  v_i - c_k \right\Vert^2, \forall k }\}
\end{equation}

(ii) After assignment, the center locations $\{c_1 .. c_K \}$ are set again to be the average feature in each group, which minimizes the WCSS (Eq.\ref{eq:wcss}) loss without the constraint. Precisely, we recompute each group center according to the image assignment $S_k$: $c_j= \frac{1}{| S_j |} \sum_{v \in S_j} v$. However, this is an infeasible solution as group centers will generally not be in ${\mathcal{W}}_q$. (iii) We therefore replace each group center $c_j$ with its nearest neighbor concept in $p_j \in W_q$. The result of this step is a new set of $K$ principal concepts $p_1..p_K$ that form the group centers.  Similarly to the swap in the Local Search algorithm, we only use the new centers if they obtain a smaller loss function. If no loss decreasing swap is found, we terminate.

\begin{table}[t]
\caption{Path similarity of retrieved word and the ground truth class names }
\label{tab:path}
\small
\begin{center}
\begin{tabular}{l  c c c c c c c c c c c c c c c c c c c}
\toprule
& & \multicolumn{1}{c}{CIFAR-10} & \multicolumn{1}{c}{CIFAR-20} & \multicolumn{1}{c}{STL-10}  & 	\\
\midrule
\multicolumn{2}{l}{ZS-Naive} &	3.90  &	4.13  &	4.72 \\			
\midrule
\multicolumn{2}{l}{Ours	}   &	\textbf{6.07}  &	\textbf{4.36}  &	\textbf{5.88} \\			
\bottomrule
\end{tabular}
\end{center}
\end{table}

\begin{table*}[t]
\caption{Grouping of Standard Classification Datasets (\%) }
\label{tab:cifar}

\small

\begin{center}
\begin{tabular}{l  c c c c c c c c c c c c c c c c c c c}

\toprule
& & \multicolumn{3}{c}{CIFAR-10} & \multicolumn{3}{c}{CIFAR-20} & \multicolumn{3}{c}{STL-10}  & \\  \cmidrule(lr){3-5} \cmidrule(lr){6-8} \cmidrule(lr){9-11}\cmidrule(lr){12-14}
&	&		ACC	&	NMI	&	ARI	&	ACC	&	NMI	&	ARI	&	ACC	&	NMI	&	ARI 	\\
\midrule																					
\multicolumn{2}{l}{	PT Only	} &	73.4	&	66.9	&	56.5 &	41.0	&	45.2	&	24.0 & 92.1	&	89.4	&	85.5 \\			
\multicolumn{2}{l}{	 PT+SCAN	} &	87.6	&	78.7	&	75.8
&	46.7	&	45.8	&	40.7 &
\textbf{98.3}	&	\textbf{95.8}	&	\textbf{96.4} \\		
\multicolumn{2}{l}{ZS-Naive}	 &	49.9	&	48.5	&	26.0
&	20.8	&	25.0	&	4.3 &
56.6	&	56.9	&	40.2 \\			
\midrule																					
\multicolumn{2}{l}{Ours	} &	\textbf{93.4}	&	\textbf{85.9}	&	\textbf{86.1} &	\textbf{48.4}	&	\textbf{51.5}	&	\textbf{34.3} & 97.9	&	95.2	&	95.4 \\

\bottomrule
\end{tabular}
\end{center}
\end{table*}

\begin{table*}[t]
\caption{Grouping of Attribute Classification Datasets (\%) }
\label{tab:non_object}

\small

\begin{center}
\begin{tabular}{l  c c c c c c c c c c c c c c c c c c c}

\toprule
& & \multicolumn{3}{c}{Stanford Activity} & \multicolumn{3}{c}{Imagenet-Dogs} & \multicolumn{3}{c}{All-Age-Faces}  & \multicolumn{3}{c}{PPMI}\\  \cmidrule(lr){3-5} \cmidrule(lr){6-8} \cmidrule(lr){9-11}\cmidrule(lr){12-14}
&	&		ACC	&	NMI	&	ARI	&	ACC	&	NMI	&	ARI	&	ACC	&	NMI	&	ARI & ACC &	NMI	& ARI	\\
\midrule																					
\multicolumn{2}{l}{	PT Only	}	&	61.4	&	66.4	&	49.0  
&	36.5	&	38.0	&	20.3
&	47.5	&	28.6	&	19.7
&	34.2	&	26.5	&	15.8\\
\multicolumn{2}{l}{	 PT+SCAN	}	&	54.0	&	66.0	&	45.3	&	49.3	&	54.4	&	36.9
&	48.8	&	33.4	&	20.2
&	27.5	&	24.1	&	12.3
\\
\multicolumn{2}{l}{ZS-Naive}		&	49.8	&	58.8	&	35.6	
&	60.9	&	65.1	&	47.8
&	50.6	&	38.7	&	24.2
&	37.9	&	38.7	&	19.2 \\
\midrule																					
\multicolumn{2}{l}{Ours	}	&	\textbf{64.9}	&	\textbf{70.6}	&	\textbf{56.4} 
&	\textbf{69.1}	&	\textbf{73.4}	&	\textbf{59.2} &
\textbf{55.4}	&	\textbf{40.6}	&	\textbf{27.4}
&	\textbf{49.0}	&	\textbf{45.1}	&	\textbf{30.2} \\												%

\bottomrule
\end{tabular}
\end{center}
\end{table*}

\textbf{Empty and excessively large group.}
In some cases, the discrete constraint results in a large proportion of images assigned to a single concept $p_k$, or one of our groups $S_k$  being empty of samples. In the case of an empty group, we replace the center location with that of a concept which would attract most samples. Specifically, we choose the concept from ${\mathcal{W}}_q$ that has the most samples as its nearest neighbors (among concepts not already in use). We also wish to address the problem of excessively large groups, that contain more than twice the average number of samples. In that case, we assign the samples in that group among all the concepts in ${\mathcal{W}}_q$ (by distance). We find the concept embedding that was chosen by the largest number of images from that excessively large group, and assign it as the new group center. Images are than reassigned between the new $K$ concepts.

\textbf{Initialization.} 
We initialize the group assignments using Ward's clustering on the image embeddings $v_1..v_{N_I}$.

\section{Experiments}
\label{sec:exp}

In this section, we evaluate our method on several grouping tasks. We demonstrate that: (i) our method can identify concepts that are closely aligned with the ground truth names (ii) it can achieve high grouping accuracy. 

\noindent  \textbf{Standard coarse-object-category datasets.} To demonstrate the effectiveness of our method with generic concept-banks, we evaluate our method on $3$ standard datasets featuring different coarse-grained objects. We use \textit{Cifar10 \cite{krizhevsky2009learning}},\textit{Cifar20 \cite{krizhevsky2009learning}}, and \textit{STL-10 \cite{coates2011analysis}}. %

\noindent  \textbf{Special attribute datasets.} We hypothesize that the concept-bank guidance is critical when the grouping attribute is  not the single coarse-grained category of the largest object in the image. To evaluate this, we performed experiments on the following datasets:
\textit{Stanford Activity \cite{yao2011human}.} A dataset presenting people performing $40$ different activities (`fixing a bike', `fixing a car', `riding a horse', etc...). The images are of very high intra-class variability (see Fig.\ref{fig:stan_act}).
\textit{People Playing Musical Instrument (PPMI) \cite{yao2010grouplet}.} A dataset of people interacting with $12$ different musical instruments (we use the \textit{PPMI+} version). %
\textit{Imagenet-Dog dataset \cite{deng2009imagenet}}
A subset of $15$ classes of the Imagenet dataset, featuring different fine-grained dogs species.
\textit{All-Age-Faces dataset \cite{cheng2019exploiting}.} A dataset containing over $82K$ images of human faces. The ground truth annotation of each person's age (between 2 and 80) is available. For evaluation, we split that dataset to five age groups in even intervals. %

\noindent  \textbf{Concept-banks.} For all the \textit{standard object datasets} we use all nouns in the WordNet dataset. This demonstrates that a large generic concept-bank can be used successfully for multiple datasets. For the \textit{special attribute datasets} we retrieve concept-banks from a ``Vocabulary Word Lists'' online resource \cite{enchanted2008wor}. We use the ``Verbs'' list for the \textit{Stanford Activity} data, the ``Musical Instruments'' list from \textit{PPMI} and the ``Dogs'' list for \textit{Imagenet-Dog}. Lastly, we demonstrate that such a list can also be composed with a set of possible numerical values; for the \textit{All-Age-Faces} dataset, we provide a list of all human ages between $0$ and $90$ years. Further details can be found in the appendix.

\noindent  \textbf{Baselines.} We evaluated our method against methods representing zero-shot learning and unsupervised clustering:

\textit{ZS-Naive.} We apply CLIP zero-shot classification between the entire list of concepts and all dataset images. We reassign each image to its closest concept. We choose the $K$ concepts to which most images were assigned as the principal concepts. We also perform zero-shot classification for each image using CLIP with the $K$ principal concepts as a baseline for our image-grouping downstream task.

As further baselines to the image-grouping downstream task, we used the following methods: \textit{PT Only.} Classical Ward's clustering with the CLIP image encoder for image feature extraction but without the concept priors. \textit{PT+SCAN.} To allow adaptation of CLIP's pretrained features, we evaluate the image clustering method SCAN \cite{van2020scan} initialized with CLIP's pretrained visual features. The pretrained features are both used for selecting the neighbors in the first stage, and as the initialization of the second stage. When the original SCAN results were better, we used them instead. %

\noindent \textbf{Metrics.} We evaluate the $K$ principal concepts by computing the WordNet path similarity \cite{pedersen2004wordnet} between each retrieved concept and each ground truth concept (class name). We find the optimal assignment between the retrieved and ground truth concepts (using bipartite matching) and report the total similarity between the sets. We only evaluate datasets that use the WordNet noun concept-bank, as other concepts are often not found in a the same WordNet hierarchy. 

\noindent To evaluate the downstream grouping performance, we use standard clustering metrics: accuracy (ACC), normalized mutual information (NMI) and adjusted Rand index (ARI).

\noindent \textbf{K principal concepts results.} We can see in Tab.\ref{tab:path} that our method finds principal concepts that are significantly closer to the groundtruth than those found by the ZS-Naive baseline.  This is also visually shown for Cifar10  in Tab.\ref{table:cifar10_nouns}(tables for Stl10 and Cifar20 can be found in the supplementary). Our method generally finds appropriate concepts. ZS-Naive often chooses many concepts belonging to the same class. E.g. `chukker', `vaulting horse` and `Seattle Slew' are all associated with the class \textit{horse}, while no concept is assigned to the equally sized class \textit{frog}. With our optimized loss function, additional concepts describe new degrees of variation, instead of repeating concepts associated with ones that were already selected (e.g. \textit{horse}).

\noindent \textbf{Image grouping results.} The results presented in Tab.\ref{tab:cifar}-\ref{tab:non_object}, demonstrate the performance of our method in grouping unlabelled images. Our method is able to utilize the language guidance for better grouping performance. We achieve strong results both on the standard object datasets and on the special attribute datasets. 
While the methods which rely only on CLIP's visual features (\textit{PT Only} and \textit{PT+SCAN}) show some image grouping capabilities, we see that the concept-bank guidance provides significant improvement on top of the ``visual only'' baselines. The gap is especially significant in the attribute classification datasets, where visual only features are often misleading with respect to the ground truth groups (Tab.\ref{tab:non_object}). On the most standard smaller datasets (Tab.\ref{tab:cifar}), we find that the visual only baselines are strong as the groups are usually defined by a single coarse grained category. The long WordNet concept-bank used is especially problematic for the \textit{ZS-Naive} baseline, which utilizes the same supervision as our method. This is because a naive use of the concept-bank can select related concepts (e.g. `chukker' and `horse') splitting single categories into multiple groups. %

\section{Discussion}
\label{sec:analysis}

\begin{figure}[t]
\centering
\caption{Grouping accuracy ($\times$), Entropy ($+$) vs. filtering level (Cifar10)}
\label{fig:quantile_unsuper}
\includegraphics[scale=0.6]{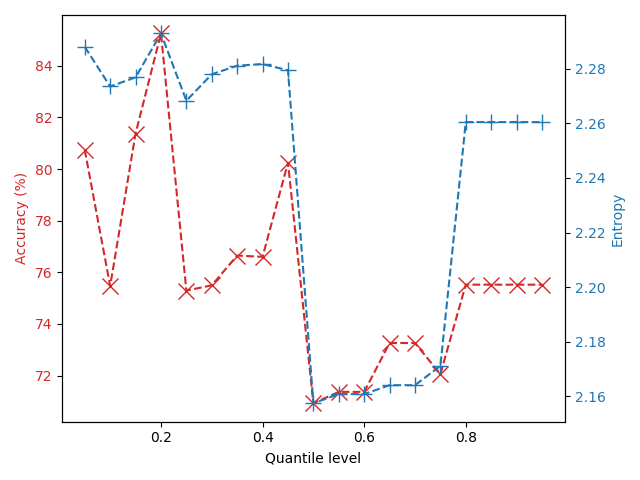}
\end{figure}

\textbf{Filtering the concept-bank.} Before running the algorithm, we filter out concepts whose ``generality'' score is above some quantile $q$, as mentioned in Sec.\ref{subsec:method:filter}. We show the effectiveness of our filtering method (without adapter) in Fig.~\ref{fig:quantile_unsuper}.

\noindent \textbf{Ground truth group name retrieval.} Our retrieved principal concepts are similar to, but not exactly the same as the groundtruth name (Tab.\ref{table:cifar10_nouns}). Our post-processing phase (Sec.\ref{sec:post_proc}) finds very similar concepts. Yet, different concepts (`milk float' \& and `Jowett') might be mapped into the same high-level concept (`vehicle').

\noindent  \textbf{Facility location optimization methods.} As explained in Sec.\ref{sec:local_relaxation}, our optimization method can be viewed as a relaxed version of the Local Search algorithm.  In the appendix we report metrics suggesting that both methods can effectively optimize the objective. 
Conversely, PAM is much slower than our method. For large dataset and larger concept-banks the time complexity of PAM is infeasible and significantly greater than that of our relaxed version.

\section{Limitations}

\noindent \textbf{Dependence on the multimodal embedding model.} Our experiments indicated that grouping using the groundtruth class names achieves strong results. Yet, our grouping performance is often limited even when we retrieve concepts very close to the true group names. It happens as the joint image-language embedding space is still not perfectly accurate. This highlights a limitation of our approach, its sensitivity to the quality of the multimodal joint embedding space. We expect this to improve as multimodal embedding methods improve.

\noindent  \textbf{No approximation guarantees of our solution to the UKFLP.} Our approach for UKFLP achieves strong results while being much faster than current methods despite its simplicity. Its main limitation is lack of theoretical guarantees. Their derivation is left for future work.

\noindent  \textbf{Increased supervision over unsupervised clustering.} Our approach requires humans to specify a single grouping attribute to derive the concept-bank. While the effort required is minor,  we argue such supervision is necessary to resolve the natural ambiguity of grouping (Fig.\ref{fig:stan_act}). %

\section{Conclusion}
We proposed a task of identifying $K$ principal components in a dataset. We reduced the task to the well-studied, uncapacitated K-facility location problem. To solve it with acceptable runtime, we suggested an efficient optimization method. Our approach is able to recover concepts very similar to the ground truth class names and accurate image grouping. We also showed that our method can provide task-specific inductive priors by concept-bank selection.

\bibliographystyle{unsrt}
\bibliography{egbib}

\clearpage

\pagenumbering{arabic}%
\renewcommand*{\thepage}{A\arabic{page}}
\begin{appendix}

\section{CLIP Adapter}
\label{subsec:method:adapter}
At the end of the optimization process aimed to find the $K$ principal concepts $\{c_1..c_k\}$ we also wish to obtain a final grouping of our image dataset.
Our discovered concepts can be used to group the dataset images into the $K$ concepts by assigning each image to its nearest concept by cosine similarity. We found that the grouping accuracy can be improved by further adaptation of the the features. Specifically, we train an adapter on top of CLIP \cite{radford2021learning} using our discovered concepts and the image grouping achieved by our method. We follow the implementation of Tip-Adapter \cite{zhang2021tip}. In this method, each image is classified by a combination of CLIP's zero-shot classification and a simple few-shot learning technique. We use Tip-Adapter with the discovered $K$ concepts, and for each of them $64$ images randomly selected from the corresponding groups $\{S_1..S_k\}$ found at the end of the WCSS optimization process.

\section{No Adapter Results:}
As an ablation, we report here the results of our method without the adapter stage using the grouping labels and $K$ principal concepts found by our method (Tab.\ref{tab:cifar_app},\ref{tab:non_object_app}). We can see that our adapter stage consistently provide an additional boost to our performance. We can further see our performance sometime approaches the performance of the ``ground-truth'' principal concepts (class names), which is an approximate upper bound for our method without the adapter stage. We report both without adapter.

\section{Implementation Details:}
\textbf{Metrics:} For the similarity measure between the retrieved $K$ principal concepts and the ground truth class names, we use the  WordNet path similarity \cite{pedersen2004wordnet}. We begin by calculating the similarity between each retrieved concepts and each class name to get an $N \times N$ matrix. We then solve the optimal assignment between the concepts and class names \cite{jonker1987shortest}. We report the sum of similarities between each concept and its assigned class name as the total similarity.

For the NMI and accuracy score we used the code\footnote{https://github.com/guysrn/mmdc/blob/main/utils/metrics.py} provided by Shiran et. al. \cite{shiran2019multi}. For the ARI score, we use the \textit{adjusted\_rand\_score} function from \textit{scikit-learn} library \cite{pedregosa2011scikit}. 

\textbf{Nearest neighbours retrieval:}
For nearest neighbours we used \textit{faiss} library \cite{johnson2019billion}.

\textbf{Grouping initialization:}
For Ward's agglomerative clustering  we use \textit{scikit-learn} library \cite{pedregosa2011scikit}.

\section{Validation:}
We designed the algorithm using the Cifar10 dataset. Parameters were kept the same for the other datasets. For the number of concpets used for abstraction ($M$) we tried two values: $\{25,50\}$. We note that the $q$ parameter (filtering level of the concept bank) is chosen for each run using an \textit{unsupervised} criterion.

Our method results does not depend on a random seed. The adapter stage does depend on a random seed, but the variation in the final results is very small ($\sim0.1\%$ accuracy).

\section{Computational Resource:}
Feature extractions should be once for each dataset and concept banks, and takes less than an hour using a GeForce RTX 2080 TI GPU.
Wards clustring can take up to on 1-hour, but can be done on a subset of the data in ~$1$ minute without degradation in the final results.
Our $K$ concepts retrieval method takes ~$1$ minute to run on each dataset (CPU only).
Using the Tip-Adapter as described takes  ~$5$ minutes using a GeForce RTX 2080 TI GPU.
All code was run using \textit{Python3} on \textit{MATE Desktop Environment}. We used \textit{numpy} version 1.19.5 and \textit{scikit-learn} version 0.24.2.

\section{Detailed Concept-Banks:}
We use the prompt \textit{"A photo showing"} to embed each of the words in each of the lists used to built the concept-bank. 

As explained in the main text, we used the WordNet nouns list for standard coarse-object-category datasets, and took all of our other lists for the special attribute datasets from a single resource. For completeness, we bring the entire lists below.

We note that our method performs well even when the list is not well curated. For example, the "Dogs" list contains words such as "bark", or "puppy", which is not specific to any single dog species.

\begin{table*}[t]
\caption{Grouping of Standard Classification Datasets (\%) }
\label{tab:cifar_app}

\small

\begin{center}
\begin{tabular}{l  c c c c c c c c c c c c c c c c c c c}

\toprule
& & \multicolumn{3}{c}{CIFAR-10} & \multicolumn{3}{c}{CIFAR-20} & \multicolumn{3}{c}{STL-10}  & \\  \cmidrule(lr){3-5} \cmidrule(lr){6-8} \cmidrule(lr){9-11}\cmidrule(lr){12-14}
&	&		ACC	&	NMI	&	ARI	&	ACC	&	NMI	&	ARI	&	ACC	&	NMI	&	ARI 	\\
\midrule

\multicolumn{2}{l}{No Adapter	}	 &	85.2  &	73.3  &	70.1	 &	40.8  &	44.5  &	 24.3  &	96.2  &	91.8  &	91.9   \\	
	
\multicolumn{2}{l}{Ours	} &	\textbf{93.4}	&	\textbf{85.9}	&	\textbf{86.1} &	\textbf{48.4}	&	\textbf{51.5}	&	\textbf{34.3} & \textbf{97.9}	&	\textbf{95.2}	&	\textbf{95.4} \\			
\midrule													\multicolumn{2}{l}{ZS-GT} &	86.7	&	75.0	&	73.1
&	54.1	&	48.8	&	32.8 &
95.9 &	91.9	&	91.4 \\

\bottomrule
\end{tabular}
\end{center}
\end{table*}

\begin{table*}[t!]
\caption{Grouping of Attribute Classification Datasets (\%) }
\label{tab:non_object_app}

\small

\begin{center}
\begin{tabular}{l  c c c c c c c c c c c c c c c c c c c}

\toprule
& & \multicolumn{3}{c}{Stanford Activity} & \multicolumn{3}{c}{Imagenet-Dogs} & \multicolumn{3}{c}{All-Age-Faces}  & \multicolumn{3}{c}{PPMI}\\  \cmidrule(lr){3-5} \cmidrule(lr){6-8} \cmidrule(lr){9-11}\cmidrule(lr){12-14}
&	&		ACC	&	NMI	&	ARI	&	ACC	&	NMI	&	ARI	&	ACC	&	NMI	&	ARI & ACC &	NMI	& ARI	\\

\midrule

\multicolumn{2}{l}{No Adapter	}	&	 63.0  & 67.9  & 52.8  &	64.5  &	65.9  &	50.8  &	\textbf{56.1}  &	38.3  &	26.5  &	\textbf{49.0}  &	41.1  &	29.6   \\

\multicolumn{2}{l}{Ours	}	&	\textbf{64.9}	&	\textbf{70.6}	&	\textbf{56.4} 
&	\textbf{69.1}	&	\textbf{73.4}	&	\textbf{59.2} &
55.4	&	\textbf{40.6}	&	\textbf{27.4}
&	\textbf{49.0}	&	\textbf{45.1}	&	\textbf{30.2} \\
\midrule												
\multicolumn{2}{l}{ZS-GT} &	82.8	&	80.1	&	72.0
&	68.4	&	70.2	&	54.8 &
60.7	&	40.8	&	30.3
&	54.5	&	44.5	&	33.5\\

\bottomrule
\end{tabular}
\end{center}
\end{table*}

\textbf{"Activities" Concept-Bank:} We used the following list: \textit{
\{accept,
ache,
acknowledge,
act,
add,
admire,
admit,
admonish,
adopt,
advise,
affirm,
afford,
agree,
ail,
alert,
allege,
allow,
allude,
amuse,
analyze,
announce,
annoy,
answer,
apologize,
appeal,
appear,
applaud,
appreciate,
approve,
argue,
arrange,
arrest,
arrive,
articulate,
ask,
assert,
assure,
attach,
attack,
attempt,
attend,
attract,
auction,
avoid,
avow,
awake,
babble,
back,
bake,
balance,
balk,
ban,
bandage,
bang,
bar,
bare,
bargain,
bark,
barrage,
barter,
baste,
bat,
bathe,
battle,
bawl,
be,
beam,
bear,
beat,
become,
befriend,
beg,
begin,
behave,
believe,
bellow,
belong,
bend,
berate,
besiege,
bestow,
bet,
bid,
bite,
bleach,
bleed,
bless,
blind,
blink,
blot,
blow,
blurt,
blush,
boast,
bob,
boil,
bolt,
bomb,
book,
bore,
borrow,
bounce,
bow,
box,
brag,
brake,
branch,
brand,
break,
breathe,
breed,
bring,
broadcast,
broil,
bruise,
brush,
bubble,
build,
bump,
burn,
burnish,
bury,
buy,
buzz,
cajole,
calculate,
call,
camp,
care,
carry,
carve,
catch,
cause,
caution,
challenge,
change,
chant,
charge,
chase,
cheat,
check,
cheer,
chew,
chide,
chip,
choke,
chomp,
choose,
chop,
claim,
clap,
clean,
clear,
climb,
clip,
close,
coach,
coil,
collect,
color,
comb,
come,
comfort,
command,
comment,
communicate,
compare,
compete,
complain,
complete,
concede,
concentrate,
concern,
conclude,
concur,
confess,
confide,
confirm,
connect,
consent,
consider,
consist,
contain,
contend,
continue,
cook,
copy,
correct,
cost,
cough,
count,
counter,
cover,
covet,
crack,
crash,
crave,
crawl,
criticize,
croak,
crochet,
cross,
cross-examine,
crowd,
crush,
cry,
cure,
curl,
curse,
curve,
cut,
cycle,
dam,
damage,
dance,
dare,
deal,
debate,
decay,
deceive,
decide,
decipher,
declare,
decorate,
delay,
delight,
deliver,
demand,
deny,
depend,
describe,
desert,
deserve,
desire,
deter,
develop,
dial,
dictate,
die,
dig,
digress,
direct,
disclose,
dislike,
dive,
divide,
divorce,
divulge,
do,
dock,
dole,
dote,
double,
doubt,
drag,
drain,
draw,
dream,
dress,
drill,
drink,
drip,
drive,
drone,
drop,
drown,
dry,
dump,
dupe,
dust,
dye,
earn,
eat,
echo,
edit,
educate,
elope,
embarrass,
emigrate,
emit,
emphasize,
employ,
empty,
enchant,
encode,
encourage,
end,
enjoin,
enjoy,
enter,
entertain,
enunciate,
envy,
equivocate,
escape,
evacuate,
evaporate,
exaggerate,
examine,
excite,
exclaim,
excuse,
exercise,
exhort,
exist,
expand,
expect,
expel,
explain,
explode,
explore,
extend,
extoll,
face,
fade,
fail,
fall,
falter,
fasten,
favor,
fax,
fear,
feed,
feel,
fence,
fetch,
fight,
file,
fill,
film,
find,
fire,
fish,
fit,
fix,
flap,
flash,
flee,
float,
flood,
floss,
flow,
flower,
fly,
fold,
follow,
fool,
force,
foretell,
forget,
forgive,
form,
found,
frame,
freeze,
fret,
frighten,
fry,
fume,
garden,
gasp,
gather,
gaze,
gel,
get,
gild,
give,
glide,
glue,
gnaw,
go,
grab,
grate,
grease,
greet,
grill,
grin,
grip,
groan,
grow,
growl,
grumble,
grunt,
guarantee,
guard,
guess,
guide,
gurgle,
gush,
hail,
hammer,
hand,
handle,
hang,
happen,
harass,
harm,
harness,
hate,
haunt,
have,
head,
heal,
heap,
hear,
heat,
help,
hide,
highlight,
hijack,
hinder,
hint,
hiss,
hit,
hold,
hook,
hoot,
hop,
hope,
hover,
howl,
hug,
hum,
hunt,
hurry,
hurt,
ice,
identify,
ignore,
imagine,
immigrate,
implore,
imply,
impress,
improve,
include,
increase,
infect,
inflate,
influence,
inform,
infuse,
inject,
injure,
inquire,
insist,
inspect,
inspire,
instruct,
intend,
interest,
interfere,
interject,
interrupt,
introduce,
invent,
invest,
invite,
iron,
irritate,
itch,
jab,
jabber,
jail,
jam,
jeer,
jest,
jog,
join,
joke,
jolt,
judge,
juggle,
jump,
keep,
kick,
kill,
kiss,
kneel,
knit,
knock,
knot,
know,
label,
lament,
land,
last,
laugh,
lay,
lead,
lean,
learn,
leave,
lecture,
lend,
let,
level,
license,
lick,
lie,
lift,
light,
lighten,
like,
list,
listen,
live,
load,
loan,
lock,
long,
look,
loosen,
lose,
love,
lower,
mail,
maintain,
make,
man,
manage,
mar,
march,
mark,
marry,
marvel,
mate,
matter,
mean,
measure,
meet,
melt,
memorize,
mend,
mention,
merge,
milk,
mine,
miss,
mix,
moan,
molt,
moor,
mourn,
move,
mow,
mug,
multiply,
mumble,
murder,
mutter,
nag,
nail,
name,
nap,
need,
nest,
nod,
note,
notice,
number,
obey,
object,
observe,
obtain,
occur,
offend,
offer,
ogle,
oil,
omit,
open,
operate,
order,
overflow,
overrun,
owe,
own,
pack,
pad,
paddle,
paint,
pant,
park,
part,
pass,
paste,
pat,
pause,
pay,
peck,
pedal,
peel,
peep,
peer,
peg,
pelt,
perform,
permit,
pester,
pet,
phone,
pick,
pinch,
pine,
place,
plan,
plant,
play,
plead,
please,
pledge,
plow,
plug,
point,
poke,
polish,
ponder,
pop,
possess,
post,
postulate,
pour,
practice,
pray,
preach,
precede,
predict,
prefer,
prepare,
present,
preserve,
press,
pretend,
prevent,
prick,
print,
proceed,
proclaim,
produce,
profess,
program,
promise,
propose,
protect,
protest,
provide,
pry,
pull,
pump,
punch,
puncture,
punish,
push,
put,
question,
quilt,
quit,
quiz,
quote,
race,
radiate,
rain,
raise,
rant,
rate,
rave,
reach,
read,
realize,
rebuff,
recall,
receive,
recite,
recognize,
recommend,
record,
reduce,
reflect,
refuse,
regret,
reign,
reiterate,
reject,
rejoice,
relate,
relax,
release,
rely,
remain,
remember,
remind,
remove,
repair,
repeat,
replace,
reply,
report,
reprimand,
reproduce,
request,
rescue,
retire,
retort,
return,
reveal,
reverse,
rhyme,
ride,
ring,
rinse,
rise,
risk,
roar,
rob,
rock,
roll,
rot,
row,
rub,
ruin,
rule,
run,
rush,
sack,
sail,
satisfy,
save,
savor,
saw,
say,
scare,
scatter,
scoff,
scold,
scoot,
scorch,
scrape,
scratch,
scream,
screech,
screw,
scribble,
seal,
search,
see,
sell,
send,
sense,
separate,
serve,
set,
settle,
sever,
sew,
shade,
shampoo,
share,
shave,
shelter,
shift,
shiver,
shock,
shoot,
shop,
shout,
show,
shriek,
shrug,
shut,
sigh,
sign,
signal,
sin,
sing,
singe,
sip,
sit,
skate,
skateboard,
sketch,
ski,
skip,
slap,
sleep,
slice,
slide,
slip,
slow,
smash,
smell,
smile,
smoke,
snap,
snarl,
snatch,
sneak,
sneer,
sneeze,
snicker,
sniff,
snoop,
snooze,
snore,
snort,
snow,
soak,
sob,
soothe,
sound,
sow,
span,
spare,
spark,
sparkle,
speak,
speculate,
spell,
spend,
spill,
spin,
spoil,
spot,
spray,
sprout,
sputter,
squash,
squeeze,
stab,
stain,
stammer,
stamp,
stand,
star,
stare,
start,
stash,
state,
stay,
steer,
step,
stipulate,
stir,
stitch,
stop,
store,
storm,
stow,
strap,
stray,
strengthen,
stress,
stretch,
strip,
stroke,
strum,
strut,
stuff,
stun,
stunt,
stutter,
submerge,
succeed,
suffer,
suggest,
suit,
supply,
support,
suppose,
surmise,
surprise,
surround,
suspect,
suspend,
sway,
swear,
swim,
swing,
switch,
swoop,
sympathize,
take,
talk,
tame,
tap,
taste,
taunt,
teach,
tear,
tease,
telephone,
tell,
tempt,
terrify,
test,
testify,
thank,
thaw,
theorize,
think,
threaten,
throw,
thunder,
tick,
tickle,
tie,
time,
tip,
tire,
toast,
toss,
touch,
tour,
tow,
trace,
track,
trade,
train,
translate,
transport,
trap,
travel,
treat,
tremble,
trick,
trickle,
trim,
trip,
trot,
trouble,
trounce,
trust,
try,
tug,
tumble,
turn,
twist,
type,
understand,
undress,
unfasten,
unite,
unlock,
unpack,
untie,
uphold,
upset,
upstage,
urge,
use,
usurp,
utter,
vacuum,
value,
vanish,
vanquish,
venture,
visit,
voice,
volunteer,
vote,
vouch,
wail,
wait,
wake,
walk,
wallow,
wander,
want,
warm,
warn,
wash,
waste,
watch,
water,
wave,
waver,
wear,
weave,
wed,
weigh,
welcome,
whimper,
whine,
whip,
whirl,
whisper,
whistle,
win,
wink,
wipe,
wish,
wobble,
wonder,
work,
worry,
wrap,
wreck,
wrestle,
wriggle,
write,
writhe,
x-ray,
yawn,
yell,
yelp,
yield,
yodel,
zip,
zoom
\}
}

\textbf{"Dogs" Concept-Bank:} We used the following list: \textit{
\{
Afghan hound,
African wild dog,
Airedale terrier,
akita,
Alaskan malamute,
American cocker spaniel,
Australian cattle dog,
bark,
basenji,
basset hound,
beagle,
bergamasco,
bichon frise,
bird dog,
bloodhound,
border collie,
borzoi,
Boston terrier,
boxer,
breed,
briard,
Brittany,
bull terrier,
bulldog,
bullmastiff,
cairn terrier,
Cape hunting dog,
chihuahua,
Chinese crested dog,
chow chow,
cocker spaniel,
collie,
companion dog,
coon hound,
corgi,
cur,
dachshund,
Dalmatian,
dhole,
dingo,
Doberman pinscher,
dog,
elkhound,
feist,
fighting dog,
fox terrier,
foxhound,
German shepherd,
golden retriever,
great Dane,
great Pyrenees,
greyhound,
growl,
guard dog,
gun dogs,
harrier,
herding dog,
hound,
hunting dog,
husky,
Irish setter,
Jack Russell terrier,
keeshond,
kerry blue terrier,
King Charles spaniel,
Labrador retriever,
lap dog,
Lhasa apso,
malamute,
Maltese,
mastiff,
Mexican hairless,
miniature schnauzer,
mongrel,
mutt,
Newfoundland,
Norfolk terrier,
old English sheepdog,
papillon,
pedigree,
pekingese,
pinscher,
pit bull,
pointer,
police dog,
Pomeranian,
poodle,
Portuguese water dog,
pug,
pup,
puppy,
purebred,
rat terrier,
rescue dog,
retriever,
Rhodesian ridgeback,
Rottweiler,
Saluki,
samoyed,
scent hound,
schnauzer,
Scottish terrier,
search-and-rescue dog,
service dog,
setter,
Siberian husky,
sighthound,
sled dog,
spaniel,
spitz,
springer spaniel,
St. Bernard,
terrier,
toy dog,
utonagan,
vizsla,
water dog,
weimaraner,
Welsh corgi,
West Highland white terrier,
Westie,
wheaten terrier,
whippet,
wild dog,
working dog,
Yorkshire terrier
\}
}

\textbf{"Musical Instruments" Concept-Bank:} We used the following list: \textit{
\{
accordion,
acoustic guitar,
Aeolian harp,
Alphorn,
alto saxophone,
anvil,
baby grand piano,
bagpipe,
balalaika,
bandoneon,
bandura,
banjo,
baritone horn,
bass,
bass clarinet,
bass drum,
bass guitar,
bassoon,
bell,
bongo drum,
bouzouki,
bow,
brass instruments,
bugle,
calliope,
carillon,
castanets,
celesta,
cello,
Celtic harp,
chimes,
cimbalom,
clarinet,
classical guitar,
clavichord,
clavier,
concertina,
conch,
conga drum,
contrabass,
cornet,
cowbell,
cymbals,
didgeridoo,
double bass,
drum,
drumsticks,
dulcimer,
electric guitar,
electric organ,
English horn,
euphonium,
fiddle,
fife,
flugelhorn,
flute,
French horn,
glockenspiel,
gong,
grand piano,
guitar,
hammered dulcimer,
harmonica,
harmonium,
harp,
harpsichord,
helicon,
horn,
hurdy-gurdy,
instrument,
jaw harp,
Jew's harp,
kazoo,
kettledrum,
keyboard,
lute,
lyre,
mallets,
mandolin,
maracas,
marimba,
mellophone,
melodeon,
Moog synthesizer,
musical instruments,
musical saw,
mute,
oboe,
ocarina,
organ,
pan pipes,
penny whistle,
percussion,
piano,
piccolo,
pipa,
pipe organ,
player piano,
pump organ,
rainstick,
rattle,
recorder,
reed,
saw,
saxophone,
sitar,
slide whistle,
snare drum,
sousaphone,
spinet,
spoons,
steel drum,
steel guitar,
string bass,
string instruments,
strings,
synthesizer,
tabla,
tambourine,
theremin,
thumb piano,
timpani,
tin whistle,
tom-tom drum,
triangle,
trombone,
trumpet,
tuba,
tubular bells,
U-V,
ukulele,
upright piano,
valve,
vibraphone,
viola,
viola da gamba,
violin,
violoncello,
vuvuzela,
Wagner tuba,
washboard,
whistle,
wind chime,
wind instruments,
woodwind instruments,
xylophone,
zither
\}
}

\textbf{"Human Ages List" Concept-Bank:} 
We used the phrase "* years old person", where * takes any integer value between 1 and 90.

\end{appendix}
\end{document}